\documentclass{article} 
\pdfoutput=1

\usepackage{nips13submit_e,times}
\usepackage{hyperref}
\usepackage{url}

\usepackage{amsmath}
\usepackage{amsfonts}
\usepackage{booktabs}
\usepackage{graphicx}
\usepackage{multirow}

\DeclareMathOperator*{\argmax}{argmax}

\title{Domain adaptation for sequence labeling \\
  using hidden Markov models}

\author{
\'Edouard Grave \\
Inria - Sierra Project-Team \\
\'Ecole Normale Sup\'erieure \\
Paris, France \\
\And
Guillaume Obozinski \\
Universit\'e Paris-Est, LIGM \\
\'Ecole des Ponts - ParisTech \\
Marne-la-Vall\'ee, France \\
\And
Francis Bach\\
Inria - Sierra Project-Team \\
\'Ecole Normale Sup\'erieure \\
Paris, France \\
}

\nipsfinalcopy 

\begin{document}

\maketitle

\vspace{-2mm}

\begin{abstract}
Most natural language processing systems based on machine learning are not robust to domain shift. For example, a state-of-the-art syntactic dependency parser trained on Wall Street Journal sentences has an absolute drop in performance of more than ten points when tested on textual data from the Web. An efficient solution to make these methods more robust to domain shift is to first learn a word representation using large amounts of unlabeled data from both domains, and then use this representation as features in a supervised learning algorithm. In this paper, we propose to use hidden Markov models to learn word representations for part-of-speech tagging. In particular, we study the influence of using data from the source, the target or both domains to learn the representation and the different ways to represent words using an HMM.
\end{abstract}

\section{Introduction}
\vspace{-1mm}
Nowadays, most competitive natural language processing systems are based on supervised machine learning. Despite the great successes obtained by those techniques, they unfortunately still suffer from important limitations. One of them is their sensitivity to domain shift: for example, a state-of-the-art part-of-speech tagger trained on the Wall Street Journal section of the Penn treebank achieves an accuracy of $97 \%$ when tested on sentences from the Wall Street Journal, but only $90 \%$ when tested on textual data from the Web~\cite{petrov2012overview}. This drop in performance can also be observed for other tasks such as syntactic parsing or named entity recognition.

One of the explanations for this drop in performance is the big lexical difference that exists accross domains. This results in a lot of out-of-vocabulary words (OOV) in the test data, \emph{i.e.}, words of the test data that were not observed in the training set. For example, more than $25\%$ of the tokens of the test data from the Web corpus~\cite{petrov2012overview} are unobserved in the training data from the WSJ. By comparison, only $11.5\%$ of the tokens of the test data from the WSJ are unobserved in the training data from the WSJ. Part-of-speech taggers make most of their errors on those out-of-vocabulary words. 

Labeling enough data to obtain a high accuracy for each new domain is not a viable solution. Indeed, it is expensive to label data for natural language processing, because it requires expert knowledge in linguistics. Thus, there is an important need for transfer learning, and more precisely for domain adaptation, in computational linguistics. A common solution consists in using large quantities of unlabeled data, from both source and target domains, in order to learn a good word representation. This representation is then used as features to train a supervised classifier that is more robust to domain shift. Depending on how much data from the source and the target domains are used, this method can be viewed as performing semi-supervised learning or domain adaptation. The goal is to reduce the impact of out-of-vocabulary words on performance. This scheme was first proposed to reduce data sparsity for named entity recognition~\cite{freitag2004trained,miller2004name}, before being applied to domain adaptation for part-of-speech tagging~\cite{huang2009distributional,huang2011language} or syntactic parsing~\cite{candito2011word,seddah2012alpage,hayashi2012naist,wu2012semi}.

Hidden Markov models have already been considered in previous work to learn word representations for domain adaptation~\cite{huang2009distributional,huang2011language} or semi-supervised learning~\cite{grave2013hidden}. Our contributions in this paper are mostly experimental: we compare different word representations that can be obtained from an HMM and study the effect of training the unsupervised HMM on source, target or both domains. While previous work mostly use Viterbi decoding to obtain word representations from an HMM, we empirically show that posterior distributions over latent classes give better results.

\section{Method}
\vspace{-1mm}
In this section, we describe the method we use to perform domain adaptation for word sequence labeling. We suppose that we have large quantities of unlabeled word sequences from both the source domain and the target domain. In addition, we suppose that we have labeled sentences only for the source domain. Our method is a two-step procedure: first, we train an unsupervised model on the unlabeled data, second, we use this model to obtain word representations used within a supervised classifier trained on labeled data. We now describe in greater details each step of the method.

\subsection{First step: unsupervised learning of an HMM}
\vspace{-1mm}
We start by training a hidden Markov chain model using unlabeled data. The words forming a sequence are denoted by the $K$-tuple $\mathbf{w} = (w_1, ..., w_K) \in \{ 1, ..., V\}^K$, where $K$ is the length of the sequence and $V$ is the size of the vocabulary. Similarly, we note the corresponding latent classes by the $K$-tuple $\mathbf{c} = (c_1, ..., c_K) \in \{ 1, ..., N\}^K$, where $N$ is the number of latent classes. The joint probability distribution on words and classes thus factorizes as
\begin{equation*}
p(W = \mathbf{w},\ C = \mathbf{c}) = \prod_{k=1}^K p(C_k = c_k \ | \ C_{k-1} = c_{k-1}) \ p(W_k = w_k \ | \ C_k = c_k).
\end{equation*}
We follow the method described by~\cite{grave2013hidden}, which consists in using the online variant of the expectation-maximization algorithm proposed by~\cite{cappe2009online} in order to scale to datasets with millions of word sequences, and to use an approximate message passing algorithm~\cite{pal2006sparse} to perform inference. Its complexity is $O(N \log N)$, where $N$ is the number of latent classes, compared to the $O(N^2)$ complexity of the exact message passing algorithm.\footnote{We use the \emph{k-best} approximation described by \cite{grave2013hidden}, setting $k$ to $3 \log_2(N)$.} This allows us to train models with hundreds of latent classes on tens of millions of sequences on a single core in a day.

\subsection{Second step: supervised learning of a CRF}
\vspace{-1mm}
\label{wordrep}
We then use the learnt unsupervised hidden Markov model to obtain features that are used to train a conditional random field~\cite{lafferty2001conditional} using the labeled sequences. We consider three kinds of features that can be obtained using an HMM. The first one, referred to as \textsc{Viterbi}, consists in computing the most probable sequence of latent classes corresponding to the sentence $\mathbf{w}$ by using Viterbi decoding:
\begin{equation*}
\mathbf{\hat c} = \argmax_{\mathbf{c}} p(C = \mathbf{c} \ | \ W = \mathbf{w}) = \argmax_{\mathbf{c}} p(W = \mathbf{w},\ C = \mathbf{c}).
\end{equation*}
Then we use the latent class $\mathbf{\hat c}_k$ as a categorical feature for the $k$th word of the sequence. The second one, referred to as \textsc{Posterior-Token}, consists in computing for the $k$th word of the sentence~$\mathbf{w}$ its posterior distribution $\mathbf{u}^{(k)}$ over the $N$ latent classes:
\begin{equation*}
u_{c}^{(k)} = \mathbb{E} \left[ \mathbf{1}{\{C_k = c\}} \ | \ W = \mathbf{w} \right].
\end{equation*}
This distribution is then used as $N$ continuous features. Finally, the last one, referred to as \textsc{Posterior-Type}, consists in computing for each word type $\tilde w$ of the vocabulary its posterior distribution $\mathbf{v}^{(\tilde w)}$ over the $N$ latent classes, averaged over all the utterances of this word type in the unlabeled data $\mathcal{U}$:
\begin{equation*}
\mathbf{v}^{(\tilde w)} = \frac{1}{Z_{\tilde w}} \sum_{i \in \mathcal{U} \ : \ w_i = \tilde w} \mathbf{u}^{(i)}
\end{equation*}
where $Z_{\tilde w}$ is the number of utterances of the word type $\tilde w$.

Let us briefly discuss the differences between those three word representations. First, the \textsc{Viterbi} representation is categorical, while both \textsc{Posterior} representations are vectorial and continuous. They thus capture much finer information, but on the other hand, lead to a slower learning algorithm for CRF, since they are not sparse. Another difference is the fact that both the \textsc{Viterbi} and \textsc{Posterior-Token} representations depend on the context, while the \textsc{Posterior-Type} representation is independent of the context, and thus all tokens with the same word type have the same representation. \textsc{Viterbi}, \textsc{Posterior-Token} and \textsc{Posterior-Type} representations have been previously used for  weakly supervised learning (see \cite{huang2009distributional,huang2011language,grave2013hidden}), but were never compared.

\section{Experiments}
\vspace{-1mm}
In this section, we describe the experiments we carried out on part-of-speech tagging. We use the universal part-of-speech tagset, introduced by~\cite{petrov2012universal}, which comprises twelve coarse tags. As a baseline, we use a CRF trained on the source domain, without any word representation or other features but the word type. In the following, we consider HMM with $128$ latent classes.

\subsection{Datasets}
\vspace{-1mm}
In all our experiments, the source domain corresponds to news articles. We use the $2,000$ first sentences of the Wall Street Journal section of the Penn treebank as labeled data for training the CRF. The first ten years of the New York Times corpus (1987-1997) are used as unlabeled data to train the hidden Markov model. This corresponds to approximately $15$ millions sentences and $300$ millions tokens.

The first target domain we consider corresponds to textual data from abstracts of biomedical articles. We use the first $5,000$ sentences of the Genia treebank~\cite{tateisi2005syntax} to evaluate our method. In this test set, approximately $40\%$ of the tokens were not observed in the training set. For the unlabeled data used to train the HMM, we have collected approximately $8$ millions sentences using the PubMed online library,\footnote{http://www.ncbi.nlm.nih.gov/pubmed} by performing a search with the keyword \emph{cancer}.

The second target domain we consider corresponds to textual data from Twitter. We use the $547$ sentences of the \textsc{Daily547} test set of~\cite{owoputi2013improved} to evaluate our method.  In this test set, approximately $35\%$ of the tokens were not observed in the training set. For the unlabeled data used to train the HMM, we have collected approximately 2 millions Twitter messages in English.

\begin{table}
\centering
\begin{tabular}{lcccccccc}
\toprule
& \hspace{1mm} & \multicolumn{3}{c}{\textbf{Biomedical}}
& \hspace{2mm} & \multicolumn{3}{c}{\textbf{Twitter}} \\
\midrule
&& \textsc{Source} & \textsc{Both} & \textsc{Target} 
&& \textsc{Source} & \textsc{Both} & \textsc{Target} \\
\midrule
\textsc{Viterbi}         && 87.8 & 90.2 & 89.9 && 64.5 & 66.5 & 67.4 \\
\textsc{Posterior-Token} && 90.4 & 91.5 & 91.4 && 67.6 & 68.5 & 70.4 \\
\textsc{Posterior-Type}  && 90.4 & 89.9 & 92.2 && 68.5 & 69.4 & 71.5 \\
\textsc{Posterior-Both}  && 91.7 & 92.0 & \textbf{92.9} && 69.4 & 70.3 & \textbf{72.4} \\
\midrule
\textsc{Baseline}        &&      & 84.6 & && & 62.4 \\
\bottomrule
\end{tabular}
\caption{Domain adaptation results for part-of-speech tagging. Unsupervised models were trained using sentences only from the news domain (\textsc{Source}), from both domains (\textsc{Both}) and only from the target domain (\textsc{Target}). The labeled data used to train the CRF \emph{only} come from the \emph{source} domain.}
\label{results-wordrep-bio}
\end{table}

\subsection{Comparison of word representations}
\vspace{-1mm}
In this section, we perform experiments to compare the different word representations we have introduced in section~\ref{wordrep}. In particular, we compare the three representations, \textsc{Viterbi}, \textsc{Posterior-Token} and \textsc{Posterior-Type}. We also consider a fourth representation, called \textsc{Posterior-Both}, which is just the concatenation of the two posterior representations introduced in section~\ref{wordrep}.

We also compare unsupervised HMM trained by using sentences only from the source domain, only from the target domain and from both domains, in order to determine the importance of the domain from which the unlabeled data come from. Indeed, as far as we now, in previous work, people have always considered using unlabeled data coming from both domains, and never investigated changing the weight of a sentence based on the domain from which it comes from. We recall that all the labeled sentences used to train the CRF come from the \emph{source} domain. We report results for adaptation to both domains in table~\ref{results-wordrep-bio}.

First, we observe that features obtained by Viterbi decoding are outperformed by posterior based representations. This is not really surprising, since the latter capture more information. We also note that using both posterior representations gives better results than using only one of them. Finally, the best results are attained for word representations trained only on unlabeled sentences from the target domain.

\subsection{Using labeled data from the target domain}
\begin{figure*}[t]
\centering
\includegraphics[height=6.2cm]{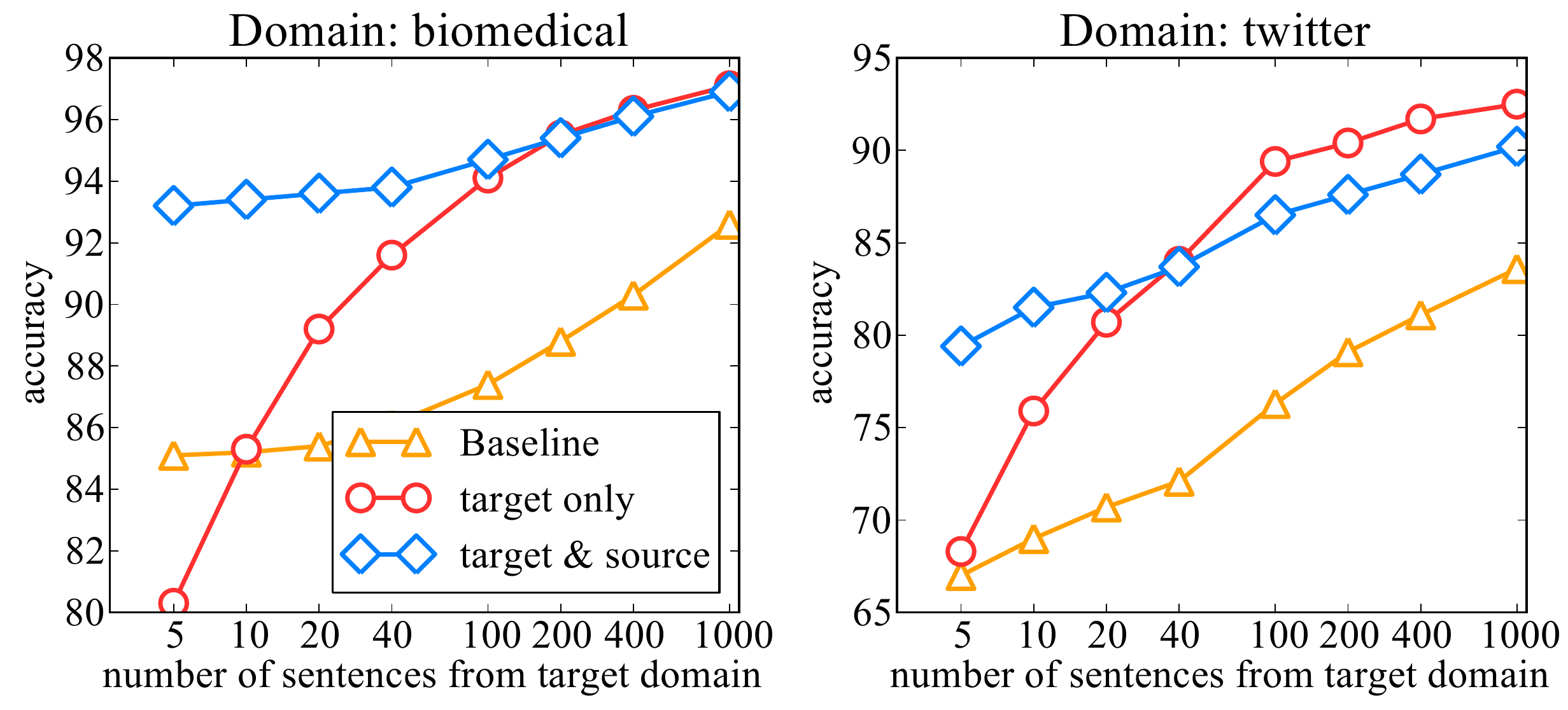}
\caption{Domain adaptation for part-of-speech tagging. In this experiment, we consider adding labeled sentences from the target domain to the training set. \textsc{Baseline} is a CRF without word representation, trained on the \emph{labeled} sentences from both domains. \textsc{Target only} is the accuracy obtained by training a CRF with word representation using only the \emph{labeled} sentences from the target domain. The scale of the $x$-axis is logarithmic.}
\label{results-target}
\end{figure*}

\vspace{-1mm}
In this section, we assume that we also have some labeled sentences from the target domain. We compare the performance obtained by training a CRF using only this small amount of labeled data with training a CRF on labeled sentences from both the source and target domains. We report results for adaptation to the biomedical domain and Twitter in figure~\ref{results-target}.

We note that adding sentences from the target domain to the training set improves the results differently for the two domains: we observe a $0.5$ point improvement for biomedical domain and $9.5$ points improvement for Twitter. Moreover we observe that when enough sentences from the target domain are available, using data from source domain for training is not useful (biomedical domain), or even gives worst results (Twitter). We have not investigated weighting sentences differently whether they come from the source or the target domain.

\section{Discussion}
\vspace{-2mm}
In this paper, we presented a very simple yet efficient method for domain adaptation for sequence tagging, and conduct experiments on part-of-speech tagging on two domains. We demonstrated that this method gives very good results when no labeled data from target domain is considered. In the case where a large enough quantity of labeled data from the target domain is used, it is not necessarily useful to use data from the source domain. Whether it is possible to still leverage data from the source domain is an interesting question for future work.

\clearpage 

\bibliography{da}{}
\bibliographystyle{unsrt}

\end{document}